\documentclass[runningheads]{llncs}

\usepackage{graphicx}
\usepackage{amssymb}
\usepackage{amsmath}
\usepackage{cite}
\usepackage{multirow}
\usepackage{marvosym}
\usepackage{color}
\usepackage{booktabs}

\begin{document}
\title{Dynamic Subframe Splitting and Spatio-Temporal Motion Entangled Sparse Attention for RGB-E Tracking
% \thanks{Supported by organization x.}
}

% \author{Anonymous submission}
\institute{}

\author{Pengcheng Shao\inst{1} \and
Tianyang Xu\inst{1} \and
Xuefeng Zhu\inst{1} \and
Xiaojun Wu\inst{1,*} \and
Josef Kittler\inst{2}
}
% \author{Hongfan Jiang\inst{1} \and
% Xiaojun Wu\inst{1} \and
% Tianyang Xu\inst{1}}
%
% \authorrunning{P.Shao et al.}
% \institute{School of Artificial Intelligence and Computer Science, Jiangnan University, Wuxi 214122, China \\
% \email{\{pengcheng.shao, xuefeng.zhu\}@stu.jiangnan.edu.cn, 
% \{tianyang.xu, wu\_xiaojun\}@jiangnan.edu.cn} \\
% Centre for Vision, Speech and Signal Processing (CVSSP), University of Surrey, Guildford GU2 TXH, UK \\
% \email{j.kittler@surrey.ac.uk}
% }

\institute{
School of Artificial Intelligence and Computer Science, Jiangnan University, Wuxi 214122,  China \\
\email{\{pengcheng.shao, xuefeng.zhu\}@stu.jiangnan.edu.cn, \{tianyang.xu, wu\_xiaojun\}@jiangnan.edu.cn}
\and Centre for Vision, Speech and Signal Processing, University of Surrey, Guildford, GU2 7XH, UK \\
\email{j.kittler@surrey.ac.uk}
}

\maketitle              % typeset the header of the contribution
\begin{abstract}
Event-based bionic camera asynchronously captures dynamic scenes with high temporal resolution and high dynamic range, offering potential for the integration of events and RGB under conditions of illumination degradation and fast motion. 
Existing RGB-E tracking methods model event characteristics utilising attention mechanism of Transformer before integrating both modalities.
Nevertheless, these methods involve aggregating the event stream into a single event frame, lacking the utilisation of the temporal information inherent in the event stream.
Moreover, the traditional attention mechanism is well-suited for dense semantic features, while the attention mechanism for sparse event features require revolution.
In this paper, we propose a dynamic event subframe splitting strategy to split the event stream into more fine-grained event clusters, aiming to capture spatio-temporal features that contain motion cues.
Based on this, we design an event-based sparse attention mechanism to enhance the interaction of event features in temporal and spatial dimensions.
The experimental results indicate that our method outperforms existing state-of-the-art methods on the FE240 and COESOT datasets, providing an effective processing manner for the event data.

\end{abstract}
\section{Introduction}\label{introduction}
Visual object tracking~\cite{xu2019joint, xu2023learning} (VOT) is a significant task in computer vision, which aims to locate the initial target in the subsequent frames.
It has been widely used in video surveillance, autonomous driving, anti-UAV~\cite{zhao2022tfatrack} and other fields~\cite{jiang2023asymmetric, xu2021adaptive}.
Most existing excellent trackers are developed based on RGB cameras. 
Owing to the impact of challenging factors in tracking scenarios, such as fast motion, background clutter and variations in light intensity, the performance of tracking algorithms still requires further enhancement.

To enhance the robustness of trackers in such challenging scenarios, researchers try to introduce new sensors~\cite{zhu2023rgbd1k, tang2023exploring,tang2024generative}.
Event camera~\cite{gallego2020event} is a novel bio-inspired camera that outputs a sparse stream of events asynchronously, capturing motion information of targets based on changes in illumination intensity.
Event camera is more sensitive to fast-moving targets due to their higher temporal resolution compared with traditional RGB cameras. 
It also works well on high speed, low power consumption and high pixel bandwidth.

\begin{figure*}[t]
\begin{center}
\includegraphics[clip,width=0.9\linewidth]{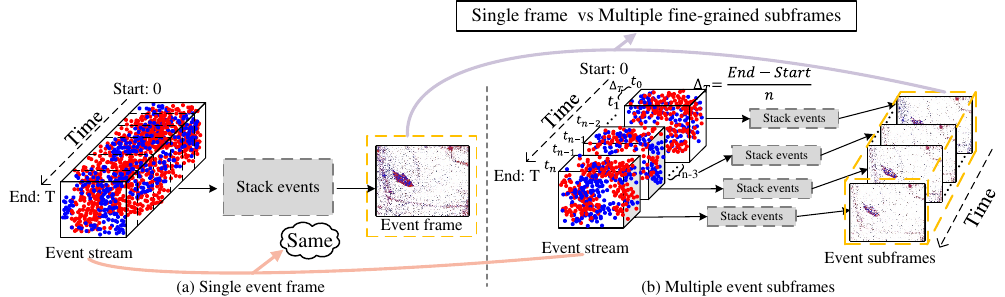}
\end{center}
\caption{
Different splitting methods for event stream. 
Blue points indicate that light intensity enhancement events occurred at that pixel at that moment.
Red points indicate that light intensity reduction events occurred at that pixel at that moment.
(a) All events in event stream are stacked together according to polarity to form a single event frame. (b) The whole event stream is divided into n smaller event streams, events in n event streams are stacked together according to polarity to form multiple event subframes.
}\label{comparison_event_frame}
\end{figure*}

Although not numerous, some studies have attempted to incorporate event datas into visual object tracking.
Zhang et al. propose STNet ~\cite{zhang2022spiking} to combine spiking neural networks and Transformer for RGB-E tracking.
Zhu et al.~\cite{zhu2023cross} utilises the Vision Transformer to bridge the distribution gap between RGB and event modalities.
While these RGB-E tracking algorithms have demonstrated commendable performance, they still face the following problems:
(1) Most RGB-E trackers generate a single event frame by summing the polarities of all events throughout the entire event stream.
However, the method of accumulating events into a single frame leads to the loss of temporal information of event pixels and fails to fully utilise the motion cues provided by the event camera.
(2) The original Transformer attention mechanism~\cite{vaswani2017attention} primarily designed for processing RGB images, encounters challenges when dealing with event frames. 
This is due to the fact that event frames mainly consist of sparse event pixels.
The attention mechanism indiscriminately processes all pixels, failing to discern between essential and irrelevant ones for tracking.
It also requires computing pairwise relationships between all pixels, which is redundant for the numerous pixels lacking motion cues related to the target.

To address the first issue that aggregating all events into a single frame, we propose a dynamic event subframe splitting strategy (DES). 
This approach partitions the event stream into $n$ small clusters enriched with temporal information, preserving the dynamic temporal information of the event modality.
As shown in Fig.~\ref{comparison_event_frame}, a single event frame represents the sum of all motion information within a specific period. 
Due to the fixedness of time slices, a single event frame is unable to encapsulate the entire integrity and continuity of motion. 
In contrast, multiple event subframes create a continuous time series, preserving the continuity of object movement. 
Simultaneously, the trajectory of a moving object can be encoded through a sequence of event frames, which is not feasible with a single event frame. 
Therefore, by utilising a series of event frames, we gain access to dynamic information that is both more accurate and complete.

To address the second issue that regarding the application of an RGB-based dense attention mechanism for event modality, we propose a spatio-temporal motion entanglement extractor (STME).
This module incorporates an event-based sparse attention (ESA) that leverages the intrinsic sparsity of event subframes.
The STME focuses on both the spatial attributes of the target at a given temporal slice and the temporal characteristics of the target across various moments, ranging from shallow to deep layers.
During this operation, STME utilises ESA to diminish unnecessary background disturbances and noisy events within the scope of the event space.
It expertly harnesses the dynamic changes of the target movement within the diverse temporal domain, thus partitioning the target motion in the scene with greater precision.

Overall, our contributions are as follows:

\begin{itemize}

\item 
We propose a dynamic event subframe splitting strategy (DES), which facilitates the capture of the motion characteristics of targets within a time period.

\item 
We propose a spatio-temporal motion entanglement extractor (STME) which incorporates an event-based sparse attention (ESA) to connect diverse event subframes for the extraction of motion information in both temporal and spatial domains.

\item 
 We introduce a novel RGB-E tracker, \textbf{D}ynamic-\textbf{S}ubframe-\textbf{M}otion-\textbf{E}ntangled-\textbf{S}parse-\textbf{A}ttention. 
Our DS-MESA surpasses the state-of-the-art trackers, showcasing the effectiveness of our approach.
\end{itemize}

\section{Related Work}
\label{relatedwork}

\subsection{Utilisation of event data in RGB-E tracking}
The integration of event cameras into visual object tracking~\cite{zhang2019visual} has gradually drawn attention in recent years.
To be more specific, DANet~\cite{fu2023distractor} utilised input event frames to propose an event-based distractor-aware tracker, comprising a motion-aware network and a target-aware network. 
It exploits motion cues and object contours provided by event frames to eliminate dynamic distractors and discover moving objects.
HRCEUTrack~\cite{zhu2023cross} employed a masking modelling strategy that randomly masks some tokens transformed from event frames, to enforce interaction between RGB and event modality. 
An orthogonal high-rank loss is proposed to mitigate network oscillations caused by the masking strategy.
AFNet~\cite{zhang2023frame} proposed a RGB-Event alignment module responsible for achieving cross-style and cross-frame-rate alignment between RGB frames and event frames, guided by motion cues provided by event frames. 
Nevertheless, these existing event-based works utilise the event modality in the form of single frame. 
It does not fully utilise the abundant information about the motion and contour details provided by the event dataset for the target.
In this paper, we discretise the event stream with finer granularity along the temporal dimension, thereby fully harnessing the temporal information of object motion conveyed by the event data.

\subsection{Attention mechanism in event-based tracking}
The Transformer attention mechanism can effectively integrate different aspects of information provided by various modalities necessary for object tracking.
VisEvent~\cite{wang2023visevent} leveraged the self-attention mechanism to enhance the correlation within the RGB modality and event modality themselves, and employed a fully-connected layer for the fusion between two modalities.
ViPT~\cite{zhu2023visual} utilised the event modality as prompts, employing the multiple Transformer encoder layers~\cite{dosovitskiy2020image} to achieve the fusion of RGB and event modalities.
Zhang et al.~\cite{zhang2023universal} proposed the global spatial dependencies extractor which employed the original self-attention mechanism to augment the features of previous event frames and used cross-attention to integrate the features of previous and current event frames.
Diverging from existing works, we introduce a sparse attention mechanism that conforms to the characteristics of sparse events, which is capable of effectively capturing the movement of the target across different subframes.

\begin{figure*}[t]
\begin{center}
\includegraphics[clip,width=1.0\linewidth]{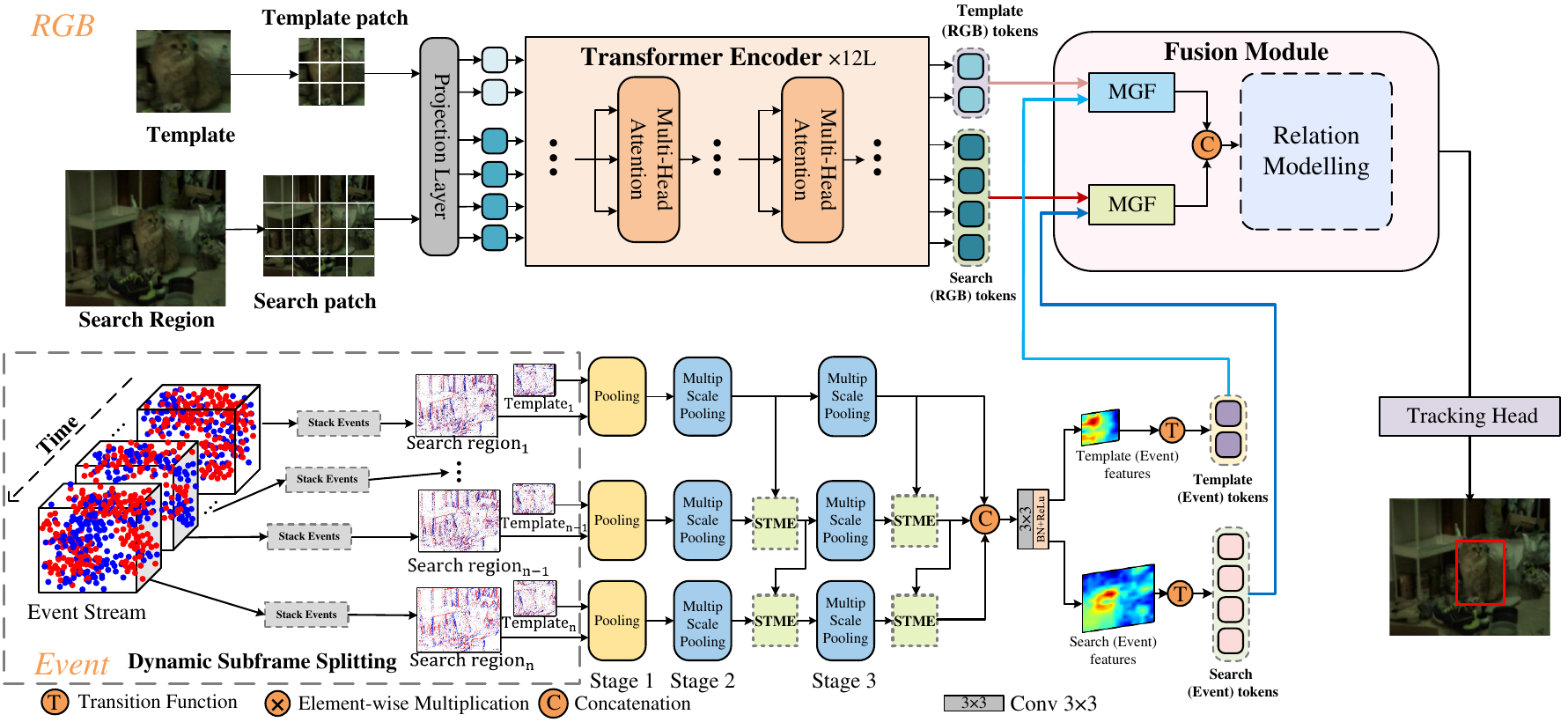}
\end{center}
\caption{
The overall architecture of DS-MESA. 
The event stream is discretised into n clusters along the temporal dimension, subsequently forming multiple event subframes. 
``STME'' stands for spatio-temporal motion entanglement module.  
``MGF'' stands for mutually guided fusion. 
The MGF employs cross-attention mechanism within both RGB and event modalities.
It comprises two layers of Transformer encoders from the Vision Transformer (ViT). 
One layer is dedicated to fusing template features across two modalities, while the other focuses on integrating search features across two modalities.
``Relation Modelling Block'' consists of N layers of Transformer encoders from ViT.
}\label{framework}
\end{figure*}

\section{Methodology}
\label{method}
\subsection{Network architecture}
The overall architecture is shown in Fig. \ref{framework}, which mainly consists of the RGB branch, the event branch, fusion module and tracking head. 
The RGB template and RGB search can be extracted by the Transformer encoder.
Dynamic event subframe splitting strategy and STME are the two core components of DS-MESA.
The goal of DES is to achieve finer-grained motion. 
While STME aims to capture motion cues from the finer-grained motion, providing spatio-temporal supplementation for the semantic features of RGB.
The fusion module effectively integrates the features extracted from the two modalities.
The tracking head predicts the bounding box of the target based on the fused search area outputted by the fusion module.
The resolution of event frames in the event datasets is 346 $\times$ 260. 
Event frames are primarily composed of blue, red and white pixels.
Blue pixels signal an increase in illumination, carrying a polarity of +1. 
Red pixels represent a decrease in illumination, with a polarity of -1. 
White pixels indicate events that have not been triggered, associated with a polarity of 0.

\noindent
\textbf{Multiple scale pooling.}
In this section, we introduce the multiple scale pooling utilised in Fig.~\ref{framework}. 
Specifically, multiple scale pooling divides the input features into four groups according to the channel dimensions.
The features of first group is operated by MaxPooling initialised with a kernel size of 3 $\times$ 3.
For the features of each subsequent group, the kernel size for MaxPooling increases by 2.
Then, we use a 1 $\times$ 1 convolution to all groups to aggregate information across groups.
Multiple scale pooling can be formulated as follows:

\begin{equation}
\begin{aligned}
& F'_E =\emph Concat[M_{3 \times 3} (\textit x_1), M_{5 \times 5}(\textit x_2),M_{7 \times 7}(\textit x_3+G_2 ), M_{9 \times 9}(\textit x_4+G_3)],
\end{aligned}
\end{equation}
\begin{equation}
F''_E={\varphi}_{1\times1}(F'_E)+F_E,
\end{equation}

\noindent
where $F_E$ is the input features of multiple scale pooling. 
$F'_E$ and $F''_E$ correspond to the features obtained during multiple scale pooling. 
The set $x=\{x_1,x_2,x_3,x_4\}$ represents the features of each division within the spatial dimension of $F_E$, which is divided into 4 groups.
 The $k_i \times k_i,$, where $ \ k_i=\{3,5,7,9\}$ denotes that the kernel size for each group is incrementally increased by 2. 
 $G_n$ denotes the results of the $n$-th group following the MaxPooling operation.

\subsection{Events representation}
Event cameras asynchronously capture the log intensity changes of each pixel, an event will be triggered when the following condition is met.

\begin{equation}
L(x, y, t) - L(x, y, t - \Delta t) \geq pC,\
\label{euqal-L}
\end{equation}

\noindent
% where \eqref{euqal-L} describes the constraints on the temporal variation of the light intensity incident on the pixel. 
where ${C}$ represents the change threshold of illumination intensity. 
${p}$ is the polarity, indicating the sign of the brightness change, where +1 and -1 correspond to positive and negative events, respectively. 
${\Delta t}$ denotes the time interval of an event occurring at the pixel position ${(x,y)}$. 
A set of events at a given time can be represented as:

\begin{equation}
E = {\{e_k}\}^{N}_{k=1} = {\{[x_k, y_k, t_k, p_k]\}}^{N}_{k=1}.
\end{equation}

\subsection{Dynamic event subframe splitting strategy (DES)}
\begin{figure*}[t]
\begin{center}
\includegraphics[clip,width=0.9\linewidth]{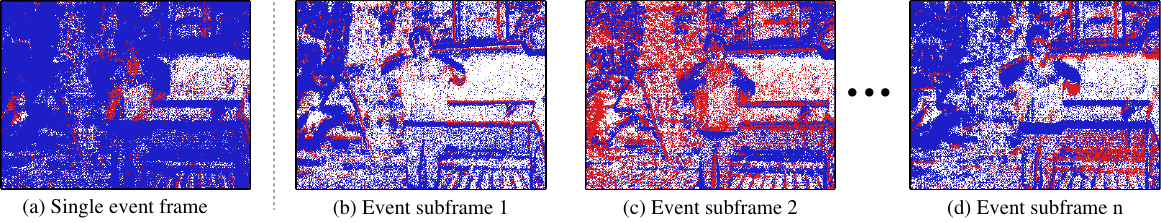}
\end{center}
\caption{Single event frame vs Multiple event subframes. (a) All events within a time interval are aggregated into a single event frame. With drastic shifts in lighting conditions, the target is almost invisible in the single event frame.
Conversely, for instances (b), (c) and (d), multiple event subframes successfully elucidate the target and its motion trajectory.
}\label{divide}
\end{figure*}

In contrast with conventional RGB cameras, which output a continuous sequence of RGB frames, event cameras necessitate the aggregation of asynchronously captured event streams into event frames based on the polarity of the events.
The widely employed method for constructing event frames includes transforming the event streams into a grid-based representation.
With frequent event occurrences, the grid-based representation fails to account for the uneven spatial distribution of event data and the temporal patterns of dense and sparse event occurrences, thereby overlooking the spatial and temporal continuity in the event data.
Therefore, we adopt an innovative dynamic subframe splitting strategy that is designed to enhance the capture of temporal resolution within event data.
We aggregate the event stream captured between the start and end exposure timestamps of an RGB frame into an n-bin voxel grid. 
% This grid is discretised into $n$ smaller temporal dimensions based on the temporal axis.
Subsequently, the grid is subdivided into $n$ smaller time dimension blocks along the temporal axis.
Following this, each ${3D}$ discretised dimension block is accumulated into a ${2D}$ frame, encapsulating the timestamps, locations of event occurrences, and the polarity of events within the pixel data.
Given an event stream $E = {\{e_k\}^{N}_{k=1}}$, $n$ event frames $F_t$ can be represented as:

\begin{equation}
\begin{aligned}
    & F_t = [ p_j \times \delta (e_j-e_k) \times pix], j\in [T + (i - 1)B, T + i B],\\
    & B = \frac{End_{rgb} - Start_{rgb}}{n}, i \in B, pix= \left\{
                                                        \begin{aligned}
                                                            & blue \ pixel, if \ p_j = 1\\
                                                            & red  \ pixel,if \ p_j = -1,\\
                                                            & white \ pixel,if \ p_j = 0
                                                        \end{aligned}
                                                        \right.
\end{aligned}
\end{equation}

\noindent
where T is the timestamp of the t-th event frame, $\delta$ is the Dirac delta function.
As shown in Fig.~\ref{divide}(a), when the target moves under rapidly changing illumination, event frames aggregated from the entire event stream fail to reflect accurately the motion characteristics of the target and are accompanied by a considerable volume of redundant event noise.
While as illustrated in Fig.~\ref{divide}(b-d), we opt to sparsify redundant events by partitioning a single event frame into multiple subframes. 
Each newly generated event subframe encompasses merely a portion of the event data, which is advantageous for emphasising the motion and fluctuation of the target.

\begin{figure*}[t]
\begin{center}
\includegraphics[clip,width=0.65\linewidth]{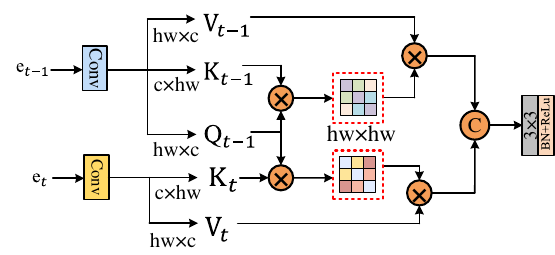}
\end{center}
\caption{A detailed architectures of the proposed STME. The red dashed boxes denote  the sparsified event attention  matrices. 
$e_{t-1}$ is derived from the features produced by the previous event frame from either stage 2 or stage 3.
$e_t$ is derived from the features produced by the previous event frame from either stage 2 or stage 3.
}\label{STME}
\end{figure*}

\begin{figure*}[t]
\begin{center}
\includegraphics[clip,width=0.8\linewidth]{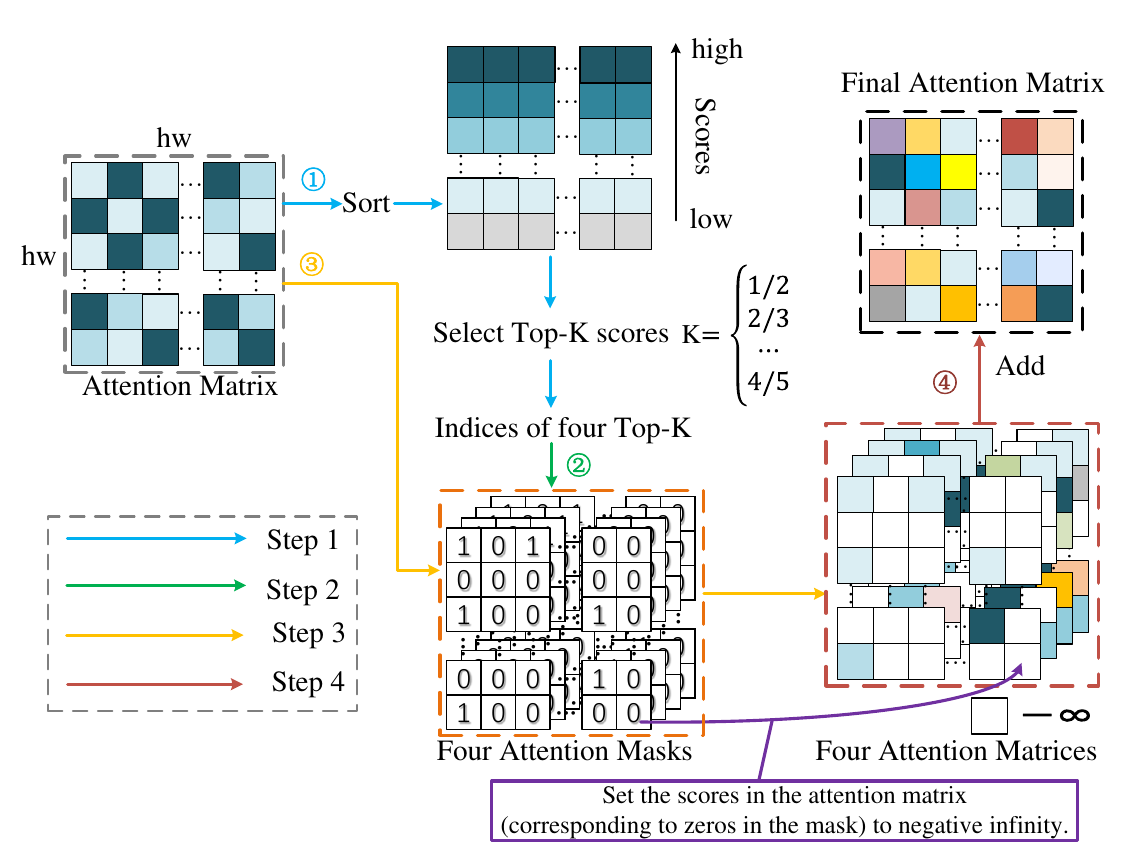}
\end{center}
\caption{
The detailed process of event-based sparse attention (ESA) operation. The obtained event attention matrix is sparsified four times and the final sparse attention matrix is acquired by adding the matrices from the four sparsifications.
}\label{fig:sparse}
\end{figure*}

\subsection{Spatio-temporal motion entanglement extractor (STME)}
As the event stream is partitioned with fine granularity, our proposed STME is aimed at capturing global spatial and temporal features containing motion cues from event data over a shorter timescale.
As depicted in Fig.~\ref{STME}, this module employs the self-attention mechanism on the previous event features $e_{t-1}$ to garner the global contextual information of the event data.
This links the motion-inducing events, thereby facilitating the capture of dynamic changes of each event across the entire space.
Since consecutive events share a similarity in their motion states, by using previous event features $e_{t-1}$ as the query for current event features $e_t$, 
attention will be guided to focus on essential information related to target motion, such as motion trajectories and morphological changes. 
Our STME can be formulated as follows:

\begin{equation}
e'_t  = \text{Concat}[\text{SparseAttn}(\frac{Q_{t-1} K_{t-1}^{\top}}{\sqrt{d_k}}) {V_{t-1}}, \text{SparseAttn}(\frac{Q_{t-1} K_t^{\top}}{\sqrt{d_k}}) {V_t}],
\end{equation}

\begin{equation}
e''_t  = \gamma (B(\phi_{3\times3} (e'_t) )),
\end{equation}

\noindent
where $e''_t$ indicates the output of the STME; 
$d_k$ is the dimension of keys;
$\phi_{k \times k}$ means a $k \times k$ convolution layer; $\gamma$ and $B$ represent the Batch Normalization and the ReLU activation function, respectively. 
$\text{SparseAttn}(\cdot)$ refer to the equation~\ref{sparse_attn}.

\subsection{Event-based sparse attention (ESA) in STME}

The traditional attention mechanism adopts a global perspective, searching and integrating information across the entire images.
Owing to the sparsity of event data, only certain regions undergoing motion contain valid information, leading to static areas or background disturbances being taken into consideration.
In order to mitigate the impact of extraneous events and background noise, we implement a sparsification process on the event attention matrix.
Within this matrix, elements with higher scores are generally indicative of relevance to the target, whereas those with lower scores tend to be associated with the background or irrelevant events.
As illustrated in Fig.~\ref{fig:sparse}, we commence by sorting the scores within the attention matrix in order of their magnitude.
Following this, dependent on the selection of K, we retain the top K scores and obtain the corresponding indices.
Subsequently, we generate a mask having all its values set to 0, with the same dimensions as the attention matrix.
Based on the acquired indices, the corresponding positions within the mask are adjusted to 1.
The scores within the attention matrix are modified in alignment with the mask:
Positions marked by a 1 in the mask retain their respective scores unchanged within the attention matrix;
positions earmarked by a 0 in the mask are assigned a value of negative infinity at the corresponding locations in the attention matrix.
Our method diverges from the fixed proportion elimination strategy in that it entails the iterative removal of the top K contributory scores from the attention matrix, this process occurring n times.
K is an adjustable parameter for dynamically controlling the level of sparsity.
Dynamic selection for attention from dense to sparse can be represented as:

\begin{equation}
\label{sparse_attn}
\text{SparseAttn}(Q,K,V) = \text{Attn} ( \lambda_1 \tau_{k_1} (\frac{Q K^{\top}}{\sqrt {d_k}})) {V}+ \dots +\lambda_n \tau_{k_n} (\frac{Q K^{\top}}{\sqrt {d_k}})) {V}),
\end{equation}

\noindent
where $\tau_{k_i}(\cdot)$ is the Top-K selection function; $\lambda_i$ is a learnable parameter used to adjust the weight distribution after undergoing different Top-K selecting.

\section{Experiments}\label{experiment}
\subsection{Implementation details}

Our DS-MESA is implemented in Python 3.7 and PyTorch 1.9.0. 
The network is trained for 60 epochs, utilising the AdamW optimiser with default settings. 
During training, we use a batch size of 16 and the initial learning rate is set $5 \times 10^{-5}$. The network training is conducted on a single RTX 3090 GPU.
The resolution of input template images is $128 \times 128$ and the search region is $256 \times 256$. The loss function of DS-MESA follows OSTrack~\cite{ye2022joint}.
To evaluate the quantitative performance of each RGB-E tracker, we employ two extensively adopted metrics: 
the precision rate (PR), which measure the centre distance between the ground truth and the predicted bounding box;
the success rate (SR), which measure the overlap between the ground truth and the predicted bounding box.

\subsection{Comparison with the state-of-the-art RGB-E Trackers}
We assess the performance of our DS-MESA using two RGB-E benchmarks: FE240~\cite{zhang2021object} and COESOT~\cite{tang2022revisiting}.
It is worth noting that both datasets were recorded using a DAVIS346 event camera with a resolution of 346 × 230 pixels. 

\noindent
\textbf{Evaluation on FE240.}
The FE240 dataset has annotation frequencies as high as 240 Hz and contains various degraded scenarios, such as high dynamic range, low light, motion blur, and fast motion. 
The dataset is partitioned into 75 training subsets and 32 testing subsets.
Tab.~\ref{tab:fe} shows the overall evaluation results on the FE240 dataset, showing a commendable performance in both precision and success rate.
Our proposed DS-MESA attains an overall precision rate (PR) of 91.9\% and a success rate (SR) of 63.8\%. 
The PR of DS-MESA is 2.7\% higher than that of DANet, and SR is 5.4\% higher than AFNet.
This demonstrates the effectiveness of our DS-MESA in fine-grained splitting of event frames and spatio-temporal motion entanglement between different event subframes.

\begin{table*}[htb]

	\caption{Overall tracking performance on FE240 and COESOT datasets. 
 }
	\centering
 % \scriptsize
 \scalebox{0.54}{
	\begin{tabular}{c|c|cccccccccc}
		\toprule
		\multirow{2}{*}{Dataset} & \multirow{2}{*}{Metrics} & \multicolumn{1}{c}{PrDiMP\cite{danelljan2020probabilistic}} & \multicolumn{1}{c}{STARKs\cite{yan2021learning}} 
        & \multicolumn{1}{c}{TransT\cite{chen2021transformer}} & \multicolumn{1}{c}{ToMP\cite{mayer2022transforming}} & \multicolumn{1}{c}{DeT\cite{yan2021depthtrack}} & \multicolumn{1}{c}{HMFT\cite{zhang2022visible}} 
        & \multicolumn{1}{c}{FENet\cite{zhang2021object}} & \multicolumn{1}{c}{AFNet\cite{zhang2023frame}}   & \multicolumn{1}{c}{DANet\cite{fu2023distractor}} & \multicolumn{1}{c}{DS-MESA}
\\
		\cmidrule{3-12} 
      	& \multicolumn{1}{c|}{} & CVPR'20 & ICCV'21  & CVPR'21	& CVPR'22  & ICCV'21 & CVPR'22 & ICCV'21  & CVPR'23  & TIP'23
   & Ours \\ 
		
		\midrule
		\multirow{2}{*}{FE240}
        &PR  & 78.3 & 79.4 & 76.2 & 83.1 & 81.2 & 84.6 & 84.3 & 87.0 & \textit{89.2} & \textbf{91.9} \\
          &SR & 51.2 & 46.2 & 49.3 & 52.3 & 54.2  & 49.1 & 55.6 & \textit{58.4}  & 56.9 & \textbf{63.8} \\
		\bottomrule
  	  % \\[pt]
  		\toprule
	\multirow{2}{*}{Dataset} & \multirow{2}{*}{Metrics} 
        & \multicolumn{1}{c}{TrDiMP\cite{wang2021transformer}} & \multicolumn{1}{c}{TransT\cite{chen2021transformer}} & \multicolumn{1}{c}{OSTrack\cite{ye2022joint}} & \multicolumn{1}{c}{AiATrack\cite{gao2022aiatrack}} 
        & \multicolumn{1}{c}{EventVOT\cite{wang2023event}} & \multicolumn{1}{c}{CEUTrack\cite{tang2022revisiting}}   & \multicolumn{1}{c}{HRCEUTrack\cite{zhu2023cross}} & \multicolumn{1}{c}{ViPT\cite{zhu2023visual}} & \multicolumn{1}{c}{TENet\cite{shao2024tenet}} & \multicolumn{1}{c}{DS-MESA} 
\\
		\cmidrule{3-12} 
      	& \multicolumn{1}{c|}{} & CVPR'21 & CVPR'21 & ECCV'22 & ECCV'22  & CVPR'24  & arXiv'22 & CVPR'23 & CVPR'23  & arXiv'24  & Ours   \\ 
		
		\midrule
		\multirow{2}{*}{COESOT}
        &PR  & 72.2 & 72.4  & 70.7 & 72.4  & 63.0  & 69.0  & 71.9 & 76.6 & \textit{76.8}  & \textbf{77.5}   \\
          &SR & 60.1  & 60.5  & 59.0 & 59.0 & 52.3  & 62.0  & 63.2 & 68.2  & \textit{68.4}  & \textbf{69.1} \\
          \bottomrule
	\end{tabular}
        }
	\label{tab:fe}
\end{table*}

% \begin{table*}[htb]
% 	\caption{State-of-the-art comparison on COESOT dataset.}
% 	\centering
%  % \scriptsize
%  \scalebox{0.59}{
% 	\begin{tabular}{c|c|cccccccccc}
% 		\toprule
% 	\multirow{2}{*}{Dataset} & \multirow{2}{*}{Metrics} 
%         & \multicolumn{1}{c}{TrDiMP\cite{wang2021transformer}} & \multicolumn{1}{c}{TransT\cite{chen2021transformer}} & \multicolumn{1}{c}{OSTrack\cite{ye2022joint}} & \multicolumn{1}{c}{AiATrack\cite{gao2022aiatrack}} 
%         & \multicolumn{1}{c}{EventVOT\cite{wang2023event}} & \multicolumn{1}{c}{CEUTrack\cite{tang2022revisiting}}   & \multicolumn{1}{c}{HRCEUTrack\cite{zhu2023cross}} & \multicolumn{1}{c}{ViPT\cite{zhu2023visual}} & \multicolumn{1}{c}{TENet} & \multicolumn{1}{c}{DS-MESA} 
% \\
% 		\cmidrule{3-12} 
%       	& \multicolumn{1}{c|}{} & CVPR'21 & CVPR'21 & ECCV'22 & ECCV'22  & CVPR'24  & arXiv & CVPR'23 & CVPR'23  & arXiv  & Ours   \\ 
		
% 		\midrule
% 		\multirow{2}{*}{COESOT}
%         &PR  & 72.2 & 72.4  & 70.7 & 72.4  & 63.0  & 69.0  & 71.9 & 76.6 & \textit{76.8}  & \textbf{77.5}   \\
%           &SR & 60.1  & 60.5  & 59.0 & 59.0 & 52.3  & 62.0  & 63.2 & 68.2  & \textit{68.4}  & \textbf{69.1} \\
% 		\bottomrule
% 	\end{tabular}
%         }
% 	\label{tab:coe}
% \end{table*}

\noindent
\textbf{Evaluation on COESOT.} 
The COESOT  dataset serves as a universally applicable single object tracking dataset, specifically designed for color event cameras. 
The dataset comprises 1354 color event videos along with 478,721 RGB frames. It is categorised into 827 training subsets and 527 testing subsets.
As shown in Tab.~\ref{tab:fe}, our proposed DS-MESA demonstrates impressive performance with a PR of 77.5\% and a SR of 69.1\%, outperforming most state-of-the-art  RGB-E trackers. This validated the effectiveness of our proposed method for event-based tracking.

\subsection{Ablation study}

\noindent
\textbf{Effectiveness of DES.} 
To validate the efficiency of DES, we compare the single event frame aggregated by the entire event stream with the multiple event subframes aggregated by the multiple event clusters. 
In our method, we split the event stream into three smaller event clusters.
The results are shown in Tab.~\ref{tab:split}, we can see that the SR and PR of ``multiple event frames" are higher than ``single event frame" on both datasets.
The results demonstrate that by splitting the event stream through DES, the tracker can effectively harness motion cues within continuous event subframes to boost its performance.

\begin{figure*}[t]
\begin{center}
\includegraphics[clip,width=1.0\linewidth]{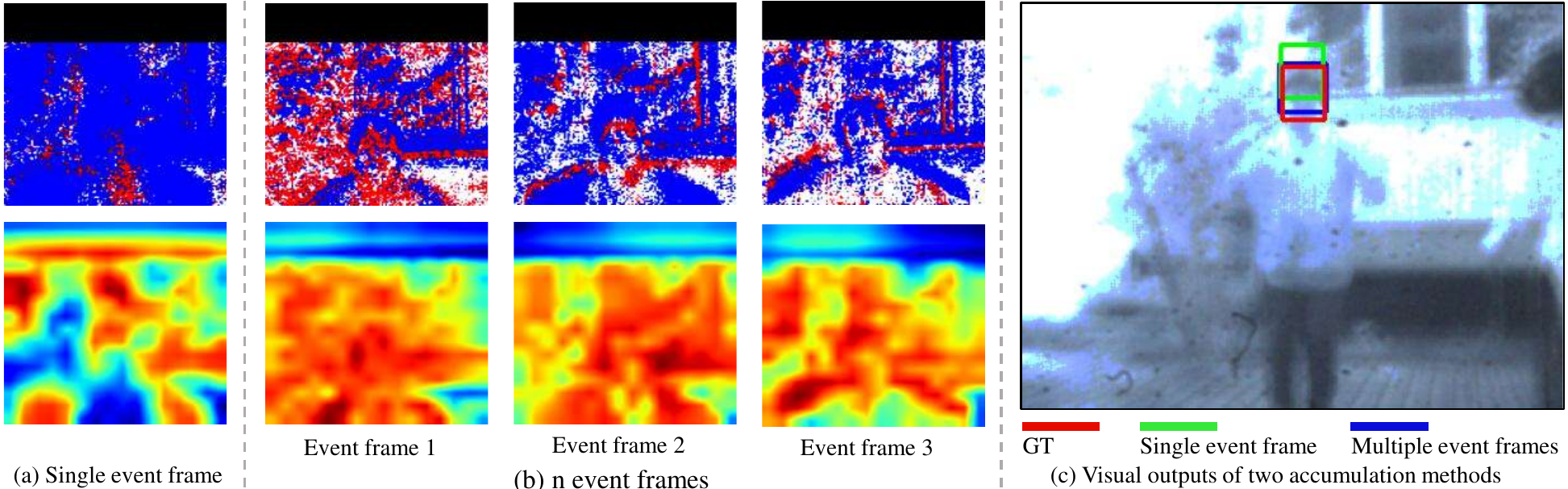}
\end{center}
\caption{
Visualisation of the comparison of two accumulation methods. In (a) and (b), the first row represents event frames and the second row represents the feature maps of event frames.
}\label{fig:visual_compare}
\end{figure*}

To further clearly demonstrate the impact of splitting the event stream on our DS-MESA, we present the visualisation of two accumulation methods in Fig.~\ref{fig:visual_compare}.
In Fig.~\ref{fig:visual_compare}(a), each pixel triggers more events when the lighting changes dramatically. 
The motion of the target is not distinctly captured in the event frame, the target area is also not highlighted in the extracted event features.
However, as illustrated in Fig.~\ref{fig:visual_compare}(b), the target area is highlighted to varying extents in the multiple event subframes.
At the same time, by comparing the bounding boxes predicted in Fig.~\ref{fig:visual_compare}(c) through two different approaches, we can clearly observe that the prediction derived from the multiple event subframes is nearer to GT.
The prediction from the single event frame is more distant from the GT, indicating that this method fails to effectively integrate and utilise the dynamic information within the event frames.

\begin{figure}
  \begin{minipage}{0.5\linewidth}
    \centering
    \includegraphics[width=1\linewidth]{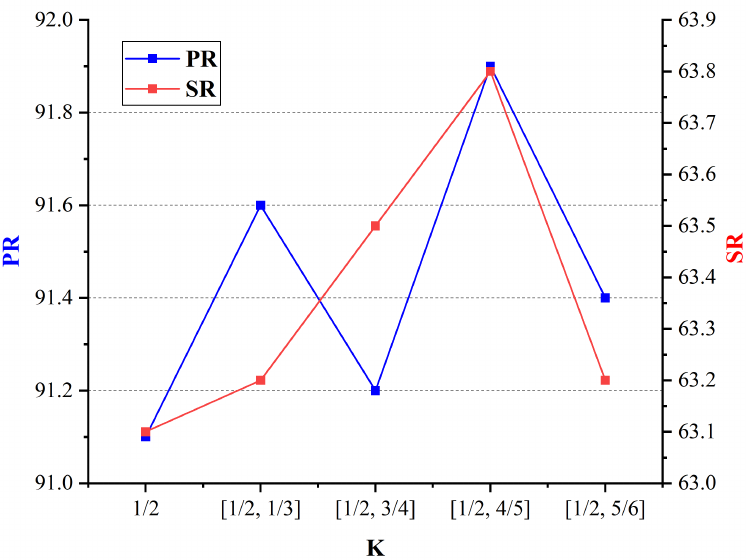}
    \caption{Ablation analysis of different k selections in STME.}
    \label{fig:diff_k}
  \end{minipage}\hfill
  \hspace{2mm}
  \begin{minipage}{0.5\linewidth}
    \centering
    \includegraphics[width=1\linewidth]{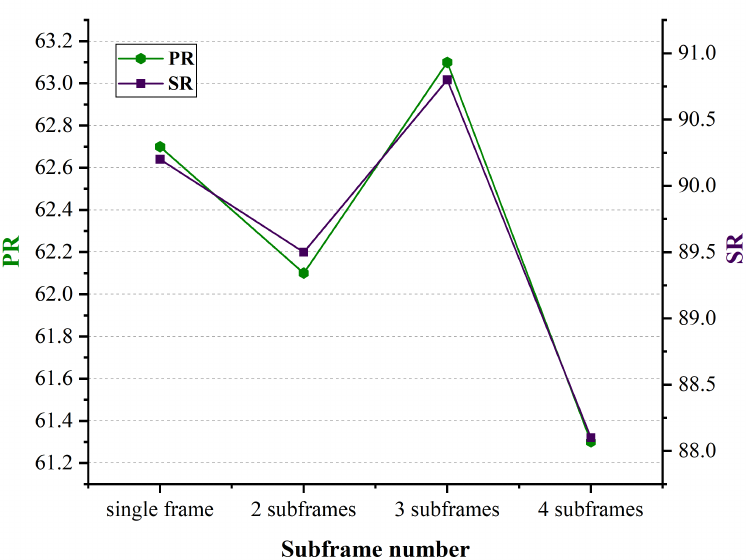}
    \caption{Comparison of different numbers of event subframes.}
      \label{fig:subframe}
  \end{minipage}\hfill

\end{figure}

\noindent
\textbf{Impact of the number of event subframes.} 
A key issue for event stream splitting is how many substreams to divide it into.
As shown in Fig.~\ref{fig:subframe}, when the event stream is partitioned into three substreams and aggregated  into three subframes, both PR and SR reach highest.
This might be attributed to the increased computational complexity caused by more substreams, and the potential omission of crucial motion information due to over-partitioning.
These results indicate that the three-subframe division  strategy significantly enhances tracker performance.

\begin{table*}[]
\caption{Comparison of single event frame and multiple event subframes.
A single event frame is accumulated from the event stream. 
$n$ event frames are accumulated from the multiple event clusters, $n=3$. 
An event stream is composed of multiple event clusters.}
\centering
\scriptsize
\begin{tabular}{c|c|cc|c|cc}
\toprule
Events accumulation methods & Dataset & PR & SR & Dataset & PR & SR \\ \midrule
single event frame & \multirow{2}{*}{FE240} & 62.7 & 90.2 & \multirow{2}{*}{COESOT} & 76.8 & 68.4 \\
n event subframes&  & 63.1 & 90.8 &  & 77.0 & 68.7 \\ \bottomrule
\end{tabular}
\label{tab:split}
\end{table*}

\noindent
\textbf{Effectiveness of STME.} 
To examine the effect of STME, we present the results of w/o STME in Tab.~\ref{tab:stme}. 
Method 1 does not implement sparse attention operations, indicating the removal of STME.
Method 5 constitutes the entire STME.
From method 1 and method 5, the results on method 1 are lower than that by using STME on method 5.
The results demonstrate that STME is beneficial in capturing and analysing spatio-temporal information within event streams, thereby enhancing the accuracy of the tracker.

\noindent
\textbf{Effectiveness of ESA.} 
The key parameter for our proposed ESA is K, with its impact illustrated in Table~\ref{tab:stme}, ranging from method 2 to method 4.
In this paper, we use 4 different Top-K combinations. ``$attn_i$ represent the sparsity rate.
If K is set to a single value, the results drop significantly. 
When K has two choices, the results show a substantial improvement.
As the number of choices for K increases to four, the results of the DS-MESA reach optimality.
By introducing K with varying sparsity rates, our DS-MESA enhances its ability to capture the dynamic changes within events more effectively.

\noindent
\textbf{Impact of different k selections in STME.} The choice of K plays a crucial role in determining the tracking performance.
As illustrated in Fig.~\ref{fig:diff_k}, we set an interval range for K in order to learn the choice with the highest scoring.
[1/2, 5/6] represents the selection of K ranging from 1/2 to 5/6, including 1/2, 2/3, 3/4, 4/5, and 5/6.
With the increase in the selection of K, the results improve gradually. 
When K falls within the range [1/2, 4/5], the optimal results are achieved.
As K increases to a certain extent, excessive subframes may lead to information overload, weakening the performance of the model.

\begin{table*}[]
\caption{Ablation study on the spatio-temporal motion entanglement extractor. ``$attn_i, i \in \{1,2,3,4\}$" indicates the different numbers of K. $attn_1 = 1/2,\ attn2 = 2/3,\ attn3 = 3/4, \ attn4 = 4/5.$}
\centering
\scriptsize
\begin{tabular}{c|cccc|c|cc|c|cc}
\toprule
\multirow{2}{*}{Method} & \multicolumn{4}{c|}{STME} & \multirow{2}{*}{Dataset} & \multirow{2}{*}{PR} & \multirow{2}{*}{SR} & \multirow{2}{*}{Dataset} & \multirow{2}{*}{PR} & \multirow{2}{*}{SR} \\ \cline{2-5} 
 & attn1 & attn2 & attn3 & attn4 &  &  &  &  &  &  \\ \midrule
1 &  &  &  &  & \multirow{5}{*}{FE240} & 90.2 & 62.7 & \multirow{5}{*}{COESOT} & 77.0 & 68.7 \\
2 & \checkmark &   &   &   &  & 91.1 & 63.1 &  & 76.1 & 67.2  \\
3 & \checkmark & \checkmark &   &   &  & 91.6 & 63.2 &  & 76.8 & 68.6 \\
4 & \checkmark & \checkmark & \checkmark &   &  & 91.2 & 63.5 &  & 77.1 & 68.9 \\
5 & \checkmark & \checkmark & \checkmark & \checkmark &  & \textbf{91.9} & \textbf{63.8} &  & \textbf{77.5} & \textbf{69.1} \\ 
\bottomrule
\end{tabular}
\label{tab:stme}
\end{table*}

\noindent
\textbf{Comparative ablation with the original Transformer attention mechanism.} 
We further investigate the impact of original dense attention. 
We replace the event-based sparse attention in the STME module with the dense attention from the ViT. 
As shown in Tab.~\ref{tab:attn}, the results of the event-based sparse attention used for STME are higher than those of the RGB-based dense attention.
This is due to the dense attention mechanism operates on each pixel within an event frame indiscriminately, failing to distinguish between noise and target. 
Consequently, this computation integrates the noise within these pixels with the target, disrupting the recognition features of the target.
In contrast, event-based sparse attention focuses on high-scoring events, filtering out the noise and enhancing the robustness of model.

\begin{table*}[]
\caption{Comparison between original attention mechanism and event-based sparse attention.
The original attention mechanism refers to the Vision Transformer (ViT).}
\centering
\scriptsize
\begin{tabular}{c|c|cc|c|cc}
\toprule
Method & Dataset & PR & SR & Dataset & PR & SR \\ \midrule
Original attention & \multirow{2}{*}{FE240} & 89.3 & 62.1 & \multirow{2}{*}{COESOT} & 76.9 & 68.5 \\
Event-based sparse attention&  & 91.9 & 63.8 &  & 77.5 & 69.1 \\ \bottomrule
\end{tabular}
\label{tab:attn}
\end{table*}

\section{Conclusion}
\label{conclusion}

In this paper, we propose a dynamic event subframe splitting strategy (DES) which divides the continuous event stream into multiple independent event clusters. 
In this way, we are able to capture the local dynamic actions of the target within specific time windows, thereby overcoming the interference posed by the integrated actions of the entire event stream.
Based on this, we also design a spatio-temporal motion entanglement module (STME) that leverages an event-based sparse attention mechanism (ESA).
The STEM focuses on the spatial distribution of the target and pays attention to continuous motion cues occurring between successive time slices.
At the same time, the ESA supports the STME in concentrating on the dynamic variations of the target within the spatio-temporal dimensions.
The extensive validation and ablation experiences show that our method has a significant performance.
Our future work will focus on splitting subframes with different granularity based on the complexity of the tracking scenes.
For simple tracking scenes, we will consider using single event frame, while for complex scenes, we will consider splitting event subframes with larger granularity in the time dimension.

\bibliographystyle{splncs04}
\bibliography{PRCV_ref}

\end{document}